\documentclass[11pt,a4paper]{article}
\usepackage[hyperref]{emnlp2020}
\usepackage{times}
\usepackage{latexsym}

\usepackage{url}

\usepackage{graphicx}
\usepackage{CJKutf8}
\usepackage{multirow}
\usepackage{array}
\usepackage{tabularx}
\usepackage{makecell}
\usepackage{amsmath} 
\usepackage{amssymb}
\usepackage{tablefootnote}
\usepackage{subcaption}
\usepackage{diagbox}
\usepackage{graphicx}
\usepackage{algorithm}
\usepackage[noend]{algpseudocode}
\usepackage{booktabs}
\usepackage{xcolor}
\usepackage{color, soul} %用color, 和 soul 包
\usepackage{colortbl}
\usepackage{makecell}
\definecolor{orange}{rgb}{1,0.5,0}
\definecolor{green}{rgb}{0,0.5,0.5}

\usepackage{microtype}

\newcolumntype{L}[1]{>{\raggedright\arraybackslash}p{#1}}
\newcolumntype{C}[1]{>{\centering\arraybackslash}p{#1}}
\newcolumntype{R}[1]{>{\raggedleft\arraybackslash}p{#1}}

\DeclareSymbolFont{extraup}{U}{zavm}{m}{n}
\DeclareMathSymbol{\varheart}{\mathalpha}{extraup}{86}
\DeclareMathSymbol{\vardiamond}{\mathalpha}{extraup}{87}

\usepackage{microtype}

\aclfinalcopy % Uncomment this line for the final submission
\def\aclpaperid{3298} %  Enter the acl Paper ID here

\title{Named Entity Recognition for Social Media Texts \\
with Semantic Augmentation
}

\author{
    Yuyang Nie$^{\diamondsuit*}$, \hspace{0.1cm}
    Yuanhe Tian$^{\varheart*}$, \hspace{0.1cm}
    Xiang Wan$^{\heartsuit}$, \hspace{0.1cm}
    Yan Song$^{\spadesuit\heartsuit\dag}$, \hspace{0.1cm}
    Bo Dai$^{\diamondsuit}$ \\
    $^{\diamondsuit}$University of Electronic Science and Technology of China\\
    $^{\varheart}$University of Washington \hspace{0.1cm} $^{\heartsuit}$Shenzhen Research Institute of Big Data\\
    $^{\spadesuit}$The Chinese University of Hong Kong (Shenzhen)\\
    \tt  $^{\diamondsuit}$nyy207@gmail.com \hspace{0.1cm} $^{\varheart}$yhtian@uw.edu \hspace{0.1cm} $^{\heartsuit}$wanxiang@sribd.cn\\
    \tt $^{\spadesuit}$songyan@cuhk.edu.cn \hspace{0.1cm} $^{\diamondsuit}$daibo@uestc.edu.cn
}

\date{}

\begin{document}
\maketitle

\def\thefootnote{*}\footnotetext{Equal contribution.}
\def\thefootnote{\dag}\footnotetext{Corresponding author.}

\def\thefootnote{\arabic{footnote}}

% idea:
% 问题：social media NER suffer from data sparsity problem
% 有一些词，有多种NE types，但其中有一些NE type
% 在整个数据集中只出现了极少次数，很容易识别错
% 
% 所以针对这个问题，我们提出了一种语义扩充的方法
% 
% 具体做法：
% 我们观察到预训练的embedding中包含丰富的语义信息，并且embedding相似的词
% 通常是语义上相似的词，且很有可能包含相同的NE type
% 1. 所以我们使用pre-trained embedding去挖掘每个词最相似的k个词
% 2. 然后采用attention的方式最这k个词做incorporation，得到最终的扩充表示
% 3. 最后再使用一个gate module把扩充表示和原来的hidden state做融合，得到最终的表示

\begin{abstract}
\textcolor{black}{
% NER方法，在短文本和非正式文本上，会遇到data sparsity这个问题
%
Existing approaches for named entity recognition suffer from data sparsity problems when conducted on short and informal texts, especially user-generated social media content.
% 语义扩充是一个好的解决方法。
%
Semantic augmentation is a potential way to alleviate this problem.
% 考虑到预训练embedding中含有丰富的语义信息，所以是一个理想的语义扩充的来源
% 
Given that rich semantic information is implicitly preserved in pre-trained word embeddings, they are potential ideal resources for semantic augmentation.
% 在这篇文章中，我们提出了一个方法，不仅考虑局部语义(来自当前句子)，还考虑扩充的语义。
%
In this paper, we propose a neural-based approach to NER for social media texts where both local (from running text) and augmented semantics are taken into account.
% 特别地，我们从大规模语料库中获取扩充的语义信息，augmentation module去编码，另外使用一个gate module融合两块信息
%
In particular, we obtain the augmented semantic information from a large-scale corpus, and propose an 
% attention-based 
attentive semantic
augmentation module and a gate module to encode and aggregate such information, respectively.
Extensive experiments are performed on three benchmark datasets collected from English and Chinese social media platforms,
where the results demonstrate the superiority of our approach to previous studies across all three datasets.\footnote{The code and the best performing models are available at \url{https://github.com/cuhksz-nlp/SANER}.}
}

\end{abstract}

\section{Introduction}

\textcolor{black}{
% 社交媒体的text通常short and informal
% 
The increasing popularity of microblogs results in a large amount of user-generated data, in which texts are usually short and informal.
How to effectively understand these texts remains a challenging task since the insights are hidden in unstructured forms of social media posts.
Thus, named entity recognition (NER) is a critical step for detecting proper entities in texts and providing support for downstream natural language processing (NLP) tasks \cite{DBLP:conf/aaai/PangLGXSC19,DBLP:conf/acl/MartinsMM19}.
}

\begin{figure}
    \centering
    \includegraphics[width=0.4\textwidth, trim=0 35 0 0]{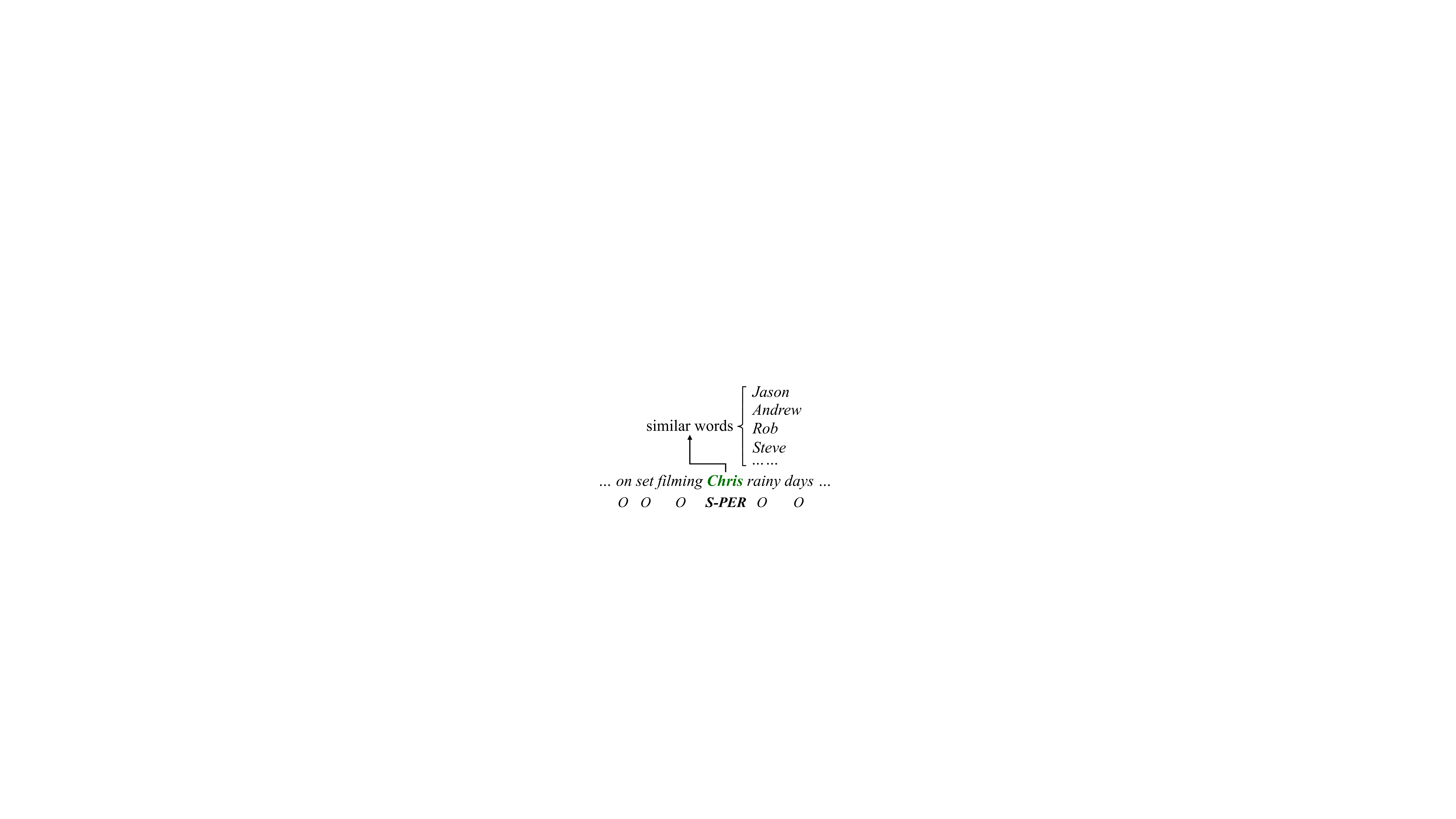}
    \caption{\textcolor{black}{An example shows that an NE tagged with ``\textit{PER}'' (Person) is suggested by its similar words.}}
    \label{fig:example}
    \vskip -1.5em
\end{figure}

However, NER in social media remains a challenging task because (i) it suffers from the data sparsity problem since entities usually represent a small part of proper names, which makes the task hard to be generalized; (ii) social media texts do not follow strict syntactic rules \cite{DBLP:conf/emnlp/RitterCME11}.
% 为了解决这些问题，之前的方法是通过 引入领域信息 (社交媒体领域训练的embedding)，以及引入额外特征 (POS，gazetteer等)来帮助提高模型在该领域上的表现。
% 具体来说，xx提出使用joint-trained embedding
% xx提出一个多任务框架，学习POS和gazetteer information
% 但是，这些方法都没有很好的解决数据稀疏性的问题。
% 我们观察到，可以引入augmented semantic information，去解决数据稀疏性的问题。
To tackle these challenges,
previous studies tired to leverage domain information (e.g., existing gazetteer and embeddings trained on large social media text) and external features (e.g., part-of-speech tags) to help with social media NER \cite{DBLP:conf/emnlp/PengD15,DBLP:conf/aclnut/AguilarMLS17}.
% 
% For instance, \citet{DBLP:conf/emnlp/PengD15} proposed to improve the model with joint-trained embeddings for Chinese social media text. 
% \citet{DBLP:conf/aclnut/AguilarMLS17} proposed a multi-task architecture to learn higher order feature representations from word and character sequences along with basic part-of-speech tags and gazetteer information. 
%
However, these approaches rely on extra efforts to obtain such extra information and suffer from noise in the resulted information.
For example, training embeddings for social media domain could bring a lot unusual expressions to the vocabulary.
% 这一篇是在cv里，用zero-shot
Inspired by studies using semantic augmentation (especially from lexical semantics) to improve model performance on many NLP tasks \cite{song-xia-2013-common,song-etal-2018-joint,kumar-etal-2019-closer,amjad-etal-2020-data}, it is also a potential promising solution to solving social media NER.
\textcolor{black}{
% Figure \ref{fig:example} shows a typical case that the term ``\textit{Chris}'', labeled with ``\textit{Person}'', only appears three times in the dataset, however, two of which are labeled with NE ``\textit{Music Artist}'', which suffers a serious data sparsity problem.
Figure \ref{fig:example} shows a typical case. ``\textit{Chris}'', supposedly tagged with ``\textit{Person}'' in this example sentence, is tagged as other labels in most cases. 
Therefore, in the predicting process, it is difficult to label ``\textit{Chris}'' correctly.
}
A sound solution is to augment the semantic space of ``\textit{Chris}'' through its similar words, such as ``\textit{Jason}'' and ``\textit{Mike}'', which can be obtained by existing pre-trained word embeddings from the general domain.

In this paper, we propose an effective approach to NER for social media texts with semantic augmentation. 
In doing so, we augment the semantic space for each token from pre-trained word embedding models, such as GloVe \cite{DBLP:conf/emnlp/PenningtonSM14} and Tencent Embedding \cite{DBLP:conf/naacl/SongSLZ18}, and encode semantic information through an 
% attention-based 
attentive semantic
augmentation module.
Then we apply a gate module to weigh the contribution of the augmentation module and context encoding module in the NER process. 
To further improve NER performance, we also attempt multiple types of pre-trained word embeddings for feature extraction, which has been demonstrated to be effective in previous studies \cite{DBLP:conf/coling/AkbikBV18,DBLP:conf/emnlp/JieL19,DBLP:conf/naacl/KasaiFFRR19,DBLP:conf/aaai/KimKK19,DBLP:journals/corr/abs-1911-04474}.
To evaluate our approach, we conduct experiments on three benchmark datasets, where the results show that our model outperforms the state-of-the-arts with clear advantage across all datasets.

\section{The Proposed Model}

\begin{figure}
    \centering
    \includegraphics[width=0.48\textwidth, trim=0 20 0 0]
    % {figure/figure_model.pdf}
    {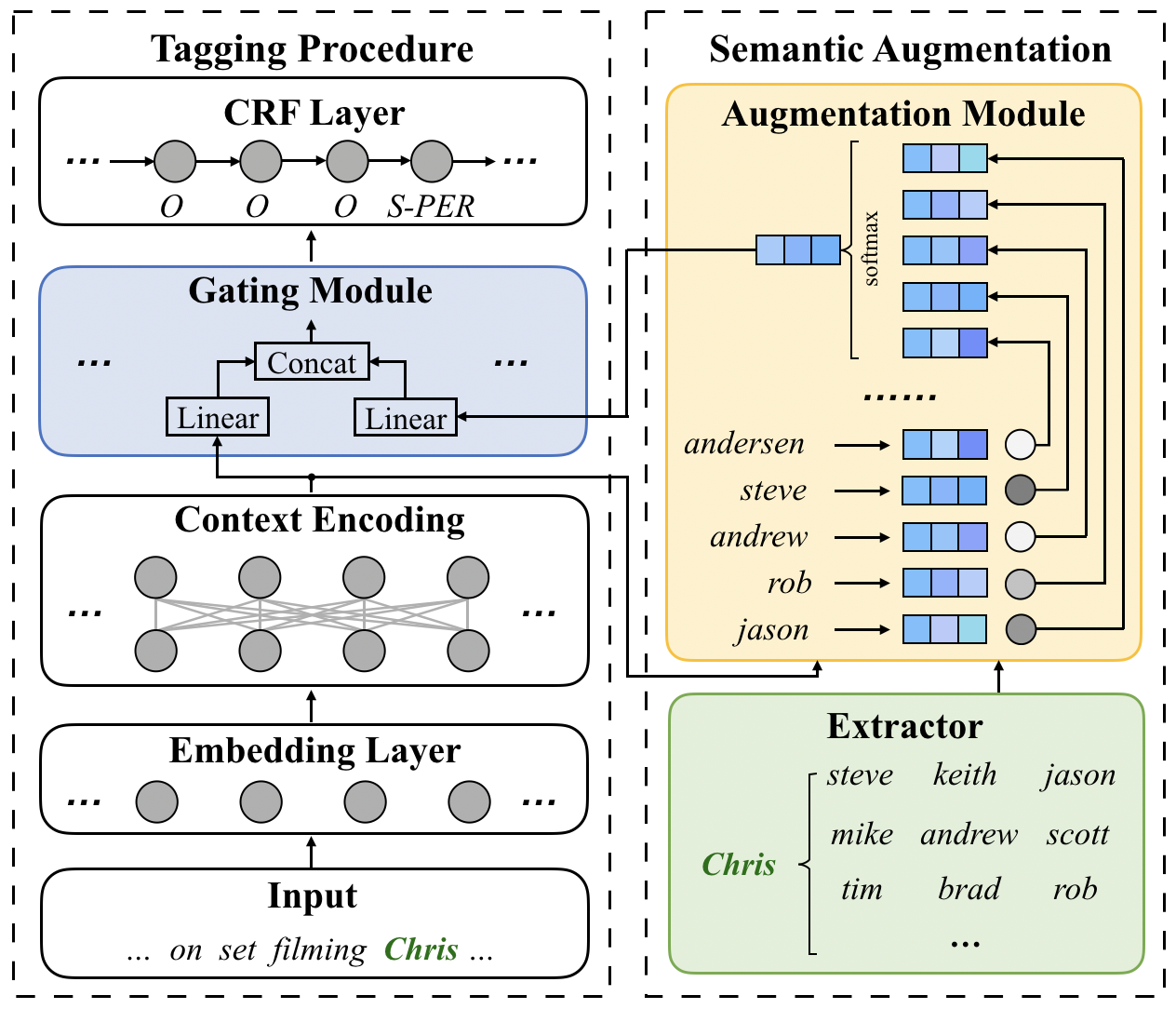}
    \caption{\textcolor{black}{
    The overall architecture of our proposed model with semantic augmentation. 
    An example sentence and its output NE labels are given, where the augmented semantic information for the word ``\textit{Chris}'' are also illustrated with the processing through the augmentation module and the gate module. 
    }}
    \label{fig:model}
    \vskip -1.5em
\end{figure}

The task of social media NER is conventionally regarded as sequence labeling task, where an input sequence $\mathcal{X} = x_1, x_2, \cdots, x_n$ with $n$ tokens is annotated with its corresponding NE labels ${\widehat{\mathcal{Y}}} = \widehat{y}_1, \widehat{y}_2, \cdots, \widehat{y}_n$ in the same length.
% 
% Following this paradigm, we propose a neural-based model illustrated in Figure \ref{fig:model} with semantic augmentation for the social media NER.
% 
% For each token in the input sentence, the semantic augmentation mechanism firstly extracts the most similar words of the token according to their pre-trained embeddings and then uses an attention mechanism to weight those word embeddings.
% Afterwards, the weighted word information is leveraged to enhance \textcolor{blue}{the backbone model} through a gate module.
% 
Following this paradigm, we propose a neural model with semantic augmentation for the social media NER.
Figure \ref{fig:model} shows the architecture of our model, where the backbone model and the semantic augmentation module are illustrated in white and yellow backgrounds, respectively.
For each token in the input sentence, we firstly extract the most similar words of the token according to their pre-trained embeddings. 
Then, the augmentation module use an attention mechanism to weight the semantic information carried by the extracted words.
Afterwards, the weighted semantic information is leveraged to enhance the backbone model through a gate module.

In the following text, we firstly introduce the encoding procedure for augmenting semantic information.
%by an 
% attention-based 
%attentive semantic augmentation module. 
Then, we present the gate module to incorporate augmented information into the backbone model.
Finally, we elaborate the tagging procedure for NER with the aforementioned enhancement. 

%
% \begin{equation}
%     \mathcal{\hat{Y}} = f(\mathcal{GA}(\mathcal{X}, \mathcal{M}(\mathcal{C})), \exists\{\mathcal{C}\} = \mathcal{E}(\mathcal{X})
% \end{equation}
% where 
% $\mathcal{C}$, obtained through the semantic extraction process (denoted as $\mathcal{E}$), refers to the context features in the augmentation module (denoted as $\mathcal{M}$). $\mathcal{GA}$ refers to the gate module to control how to use the encodings from context encoder and that from $\mathcal{M}$. 
% The details of these steps are illustrated below. 
% In the following texts, we firstly introduce how we extract the augmented semantic information, then illustrate the encoding and incorporating procedure for augmented semantic information according to augmentation module and gate module. 
% the encoding procedure for global context and local context, sequentially. Afterwards, we illustrate the incorporation of global and local context. 

% \subsection{Attention-based Encoding for Semantic Information}
\subsection{Attentive Semantic Augmentation}

The high quality of text representation is the key to obtain good model performance for many NLP tasks \cite{song2017learning,sileo-etal-2019-mining}.
However, obtaining such text representation is not easy in the social media domain because of data sparsity problem.
Motivated by this fact,
%Therefore, the motivation of the
we propose semantic augmentation mechanism for social media NER by enhancing the representation of each token in the input sentence with the most similar words in their semantic space, which can be measured by pre-trained embeddings.

In doing so, 
% we firstly obtain a lexicon with each word in it represented by a pre-trained embedding (e.g., GloVe for English and Tencent Embedding for Chinese).
% 
% Then, according to the pre-trained embeddings, 
for each token $x_i \in \mathcal{X}$,
% in the input sentence, 
we use pre-trained word embeddings (e.g., GloVe for English and Tencent Embedding for Chinese) to extract the top $m$ words that are most similar to $x_i$ based on cosine similarities and denote them as
\begin{equation}
    \setlength\abovedisplayskip{5pt}
    \setlength\belowdisplayskip{5pt}
    {C}_i = \{c_{i,1}, c_{i,2}, \cdots, c_{i,j}, \cdots, c_{i,m}\}
\end{equation}
% Before encoding augmented semantic information for the proposed model, an extractor is needed to provide such information. 
% Since pre-trained word embeddings provide substantial semantic information, a simple but effective approach for semantic augmentation is to leverage similar words for each word as its augmented semantic information, which can be formalized by:
% \begin{equation}
%     \setlength\abovedisplayskip{5pt}
%     \setlength\belowdisplayskip{5pt}
%     {C}_i = \{c_{i1}, c_{i2}, \ldots, c_{ij}, \ldots, c_{ik}\},
% \end{equation}
% where $k$ represents the top-$k$ similar words for $x_i$.
% \textcolor{blue}{
Afterwards, we use another embedding matrix to map all extracted words $c_{i,j}$ to their corresponding embeddings $\mathbf{e}_{i,j}$.
% \footnote{\textcolor{blue}{The embedding of $c_{i,j}$ does not necessarily equal to the pre-trained embedding used to compute its similarity with $x_i$}.}
% by a transmit matrix, 
% with $\mathbf{e}_{i,j}$ denoting the embedding of $c_{i,j}$.
% }
% 
% 这里加了一句attention的Motivation，即不是每一个c_ij都是有帮助的，所以我们提出一个基于attention的方法去高亮重要的信息，具体来说，...
Since not all $c_{i,j} \in C_i$ are helpful for predicting the NE label of $x_i$ in the given context, it is important to distinguish the contributions of different words to the NER task in that context.
Consider that the attention and weight based approaches are demonstrated to be effective choices to selectively leverage extra information in many tasks \cite{kumar-etal-2018-knowledge,margatina-etal-2019-attention,tian-etal-2020-joint,tian2020improving,tian-etal-2020-suppertagging,tian-etal-2020-constituency}, we propose an attentive semantic augmentation module (denoted as $AU$) to weight the words according to their contributions to the task in different contexts.
Specifically, for each token $x_i$, the augmentation module assigns a weight to each word $c_{i,j} \in C_i$ by
\begin{equation}
    \setlength\abovedisplayskip{5pt}
    \setlength\belowdisplayskip{5pt}
    p_{i,j} = \frac{exp(\mathbf{h}_i \cdot \mathbf{e}_{i,j})}{\sum_{j=i}^m exp(\mathbf{h}_i \cdot \mathbf{e}_{i,j})},
\end{equation}
where $\mathbf{h}_i$ is the hidden vector for $x_i$ obtained from the context encoder with its dimension matching that of the embedding (i.e., $\mathbf{e}_{i,j}$) of $c_{i,j}$. 
Then, we apply the weight $p_{i,j}$ to the word $c_{i,j}$ to compute the final augmented semantic representation by
\begin{equation}
    \setlength\abovedisplayskip{5pt}
    \setlength\belowdisplayskip{5pt}
    \mathbf{v}_i = \sum_{j=1}^{m} p_{i,j} \mathbf{e}_{i,j},
\end{equation}
where $\mathbf{v}_i$ is the derived output of ${AU}$, and contains the weighted semantic information.
Therefore, the augmentation module ensures that the augmented semantic information are weighted based on their contributions and important semantic information is distinguished accordingly.

\subsection{The Gate Module}

We observe that the contribution of the obtained augmented semantic information to the NER task could vary in different contexts and
a gate module (denoted by ${GA}$) is naturally desired to weight such information in the varying contexts.
Therefore, to improve the capability of NER with the semantic information, we propose a gate module to aggregate such information to the backbone NER model.
%
%Specifically, inspired by \citet{DBLP:conf/emnlp/ChoMGBBSB14} and \citet{DBLP:conf/www/ZhangNWST20},
Particularly,
we use a $reset$ gate to control the information flow by
\begin{equation}
    \setlength\abovedisplayskip{5pt}
    \setlength\belowdisplayskip{5pt}
    \mathbf{g} = \sigma(\mathbf{W}_1 \cdot \mathbf{h}_i + \mathbf{W}_2 \cdot \mathbf{v}_i + \mathbf{b}_g), 
\end{equation}
where $\mathbf{W}_1$ and $\mathbf{W}_2$ are trainable matrices and $\mathbf{b}_g$ the corresponding bias term.
Afterwards, we use
\begin{equation}
    \setlength\abovedisplayskip{5pt}
    \setlength\belowdisplayskip{5pt}
    \mathbf{u}_i = [\mathbf{g} \circ \mathbf{h}_i] \oplus [(\mathbf{1} - \mathbf{g}) \circ \mathbf{v}_i]
\end{equation}
to balance the information from context encoder and the augmentation module, where $\mathbf{u}_i$ is the derived output of the gate module;
$\circ$ represents the element-wise multiplication operation and $\mathbf{1}$ is a 1-vector with its all elements equal to $1$.

\subsection{Tagging Procedure}

\textcolor{black}{
To provide $\mathbf{h}_i$ to the augmentation module, we adopt a context encoding module (denoted as ${CE}$) proposed by \citet{DBLP:journals/corr/abs-1911-04474}.
Compared with vanilla Transformers, this encoder additionally models the direction and distance information of the input, which has been demonstrated to be useful for the NER task. 
%and achieves better performance.
% \footnote{\textcolor{orange}{The context encoding module proposed by \cite{DBLP:journals/corr/abs-1911-04474} additionally models direction and distance information of the input, which are demonstrated to be useful for NER comparing to the vanilla Transformer.}}. 
Therefore, the encoding procedure of the input text can be denoted as
\begin{equation}
    \setlength\abovedisplayskip{5pt}
    \setlength\belowdisplayskip{5pt}
    \mathbf{H} = {CE}(\mathbf{E}),
\end{equation}
where $\mathbf{H} = [\mathbf{h}_1, \mathbf{h}_2, \cdots, \mathbf{h}_n]$ and $\mathbf{E} = [\mathbf{e}_1, \mathbf{e}_2, \ldots, \mathbf{e}_n]$ are lists of hidden vectors and embeddings of $\mathcal{X}$, respectively. 
In addition, since pre-trained word embeddings contain substantial context information from large-scale corpus, and different types of them may contain diverse information, a straightforward way of incorporating them is to concatenate their embedding vectors by
\begin{equation}
    \setlength\abovedisplayskip{5pt}
    \setlength\belowdisplayskip{5pt}
    \mathbf{e}_i = \mathbf{e}_i^1 \oplus \mathbf{e}_i^2 \oplus \ldots \oplus \mathbf{e}_i^T,
\end{equation}
where $\mathbf{e}_i$ is the final word embedding for $x_i$ and ${T}$ the set of all embedding types.
Afterwards, a trainable matrix $\mathbf{W}_u$ is used to map $\mathbf{u}_i$ obtained from the gate module to the output space by $\mathbf{o}_i = \mathbf{W}_u \cdot \mathbf{u}_i$.
Finally, a conditional random field (CRF) decoder is applied to predict the labels $\widehat{y}_{i} \in {L}$ (where ${L}$ is the set with all NE labels) in the output sequence $\widehat{\mathcal{Y}}$ by
\begin{equation} \label{eq:crf}
\setlength\abovedisplayskip{5pt}
\setlength\belowdisplayskip{5pt}
    \widehat{y}_{i} = \underset{y_{i} \in {L}}{\arg \max} \frac{ exp(\mathbf{W}_{c} \cdot \mathbf{o}_{i} + \mathbf{b}_{c})}
                    {\sum_{y_{i-1}y_{i}} exp(\mathbf{W}_{c} \cdot \mathbf{o}_{i} + \mathbf{b}_{c})},
\end{equation}
where $\mathbf{W}_{c}$ and $\mathbf{b}_{c}$ are the trainable parameters to model the transition for $y_{i-1}$ to $y_{i}$.
}

\section{Experiments}

\vspace{-0.2cm}

\begin{table}[t]
    % \small
    % \begin{sc}
    \scalebox{0.84}{
    \begin{tabular}{L{1.6cm}|L{1.5cm}L{1.0cm}|R{0.73cm}R{0.85cm}R{0.85cm}}
    % }
    \toprule
    \textbf{Language}           & \textbf{Dataset}          &           & \textbf{Train} & \textbf{Dev} & \textbf{Test} \\
    \midrule
    \multirow{6}{*}{English}    & \multirow{3}{*}{W16}   & $\#$Sent. & 2,394 & 1,000 & 3,850 \\
                                &                           & $\#$Ent.  & 1,496 & 661   & 3,473 \\
                                &                           & $\%$Uns.  & -     & 52.1  & 80.0  \\
                                \cmidrule{2-6}
                                & \multirow{3}{*}{W17}   & $\#$Sent. & 3,394 & 1,008 & 1,287 \\
                                &                           & $\#$Ent.  & 1,975 & 835   & 1,079 \\
                                &                           & $\%$Uns.  & -     & 34.8  & 84.5   \\
    \midrule
    \multirow{3}{*}{Chinese}    & \multirow{3}{*}{WB}   & $\#$Sent. & 1,350 & 270 & 270 \\
                                &                           & $\#$Ent.  & 1,885 & 389   & 414 \\
                                &                           & $\%$Uns.  & -     & 51.4  & 45.2      \\
    \bottomrule
    \end{tabular}
    }
    \vspace{-0.2cm}
    \caption{The statistics of all benchmark datasets w.r.t. the number of sentences ($\#$ Sent.), named entities ($\#$ Ent.) and the percentage of unseen entities ($\%$ Uns.).
    }
    \label{tab:dataset}
    \vskip -1.5em
\end{table}

\subsection{Settings}

In our experiments, we use three social media benchmark datasets, including WNUT16 (W16) \cite{DBLP:conf/aclnut/StraussTRMX16}, WNUT17 (W17) \cite{DBLP:conf/aclnut/DerczynskiNEL17}, and Weibo (WB) \cite{DBLP:conf/emnlp/PengD15}, 
where W16 and W17 are English datasets constructed from Twitter, and WB is built from Chinese social media platform (Sina Weibo).
For all three datasets, we use their original splits and report the statistics of them in Table \ref{tab:dataset} (e.g., the number of sentences ($\#$Sent.), entities ($\#$Ent.), and the percentage of unseen entities (\%Uns.) with respect to the entities appearing in the training set).
% \textcolor{orange}{Besides, folloin the data preprocess stage, we replace all digits with ``0''. }

\begin{table*}[t]
\begin{subtable}[t]{0.45\textwidth}
    \centering
    \small
    % \begin{sc}
    \begin{tabular}{C{0.35cm}|C{0.5cm}C{0.5cm}|C{1.35cm}C{1.35cm}C{1.2cm}}
        \toprule
        \textbf{ID} & \textit{\textbf{SE}} & \textbf{\textit{GA}} & \textbf{W16} & \textbf{W17} & \textbf{WB} \\
        \midrule
        1 & $N$  & $N$          & 54.79            & 48.41            & 65.36 \\
        2 & ${DS}$& $N$         & 55.03            & 48.36            & 65.01 \\
        3 & ${DS}$&$Y$          & 56.28            & 48.98            & 66.24 \\
        4 & $AU$&$N$            & 56.86            & 49.26            & 68.21 \\
        5 & $AU$&$Y$            & \textbf{57.94}   & \textbf{50.02}   & \textbf{69.32} \\
        \bottomrule
    \end{tabular}
    % \end{sc}
    % \vspace{-0.1cm}
    \caption{Development Set}
    \label{tab:dev}
    % \vskip -1em
\end{subtable}
\hspace{0.8cm}
\begin{subtable}[t]{0.45\textwidth}
    \centering
    \small
    % \begin{sc}
    \begin{tabular}{C{0.35cm}|C{0.5cm}C{0.5cm}|C{1.35cm}C{1.35cm}C{1.2cm}}
        \toprule
        \textbf{ID} & \textit{\textbf{SE}} & \textbf{\textit{GA}} & \textbf{W16} & \textbf{W17} & \textbf{WB} \\
        \midrule
        1 & $N$  & $N$                  & 52.98        & 48.82        & 66.02 \\
        2 & ${DS}$& $N$                 & 53.11        & 48.71        & 65.78 \\
        3 & ${DS}$&$Y$                  & 54.02        & 49.56        & 67.52 \\
        4 & ${AU}$&$N$                  & 54.29        & 49.81        & 68.46 \\
        5 & ${AU}$&$Y$                  & \textbf{55.01}& \textbf{50.36}& \textbf{69.80} \\
        \bottomrule
    \end{tabular}
    % \end{sc}
    % \vspace{-0.1cm}
    \caption{\textcolor{black}{Test Set}}
    % \vskip -1em
    \label{tab:self}
\end{subtable}
\vspace{-0.2cm}
\caption{\textcolor{black}{$F1$ scores of the baseline model and ours enhanced with semantic augmentation (``${SE}$'') and the gate module (``${GA}$'') on the development (a) and test (b) sets. ``${DS}$'' and ``${AU}$'' represent the direct summation and attentive augmentation module, respectively. $Y$ and $N$ denote the use and non-use of corresponding modules.}}
\label{tab: results}
\vskip -1.5em
\end{table*}

% For model implementation, we use the \textbf{BIOES} tagging scheme instead of the standard \textbf{BIO} scheme since previous studies have shown optimistic improvement with this scheme \cite{DBLP:conf/naacl/LampleBSKD16}.
For model implementation, we follow \citet{DBLP:conf/naacl/LampleBSKD16} to use the BIOES tag schema to represent the NE labels of tokens in the input sentence.
For the text input, we try two types of embeddings for each language.\footnote{We report the results of using each individual type of embeddings in Appendix A.}
Specifically, for English, we use 
% Glove\footnote{We use the Glove.6B.100d version, downloaded from \url{https://nlp.stanford.edu/projects/glove/}.} \cite{DBLP:conf/emnlp/PenningtonSM14} , 
ELMo \cite{DBLP:conf/naacl/PetersNIGCLZ18} and BERT-cased large  \cite{DBLP:conf/naacl/DevlinCLT19};
for Chinese, we use 
% pre-trained character and bi-gram embeddings\footnote{We obtain the embeddings from \url{https://github.com/jiesutd/LatticeLSTM}. } released by \cite{DBLP:conf/acl/ZhangY18} (denoted as Giga), 
Tencent Embedding \cite{DBLP:conf/naacl/SongSLZ18}, and ZEN \cite{DBLP:journals/corr/zen}.\footnote{
We obtain the pre-trained BERT model from \url{https://github.com/google-research/bert},
Tencent Embeddings from \url{https://ai.tencent.com/ailab/nlp/embedding.html},
and ZEN from \url{https://github.com/sinovation/ZEN}.
Note that we use ZEN because it achieves better performance than BERT on different Chinese NLP tasks.
For reference, we report the results of using BERT in Appendix B.}
In the context encoding module, we use a two-layer transformer-based encoder proposed by \citet{DBLP:journals/corr/abs-1911-04474} with 128 hidden units and 12 heads.
To extract similar words carrying augmented semantic information, we use the pre-trained word embeddings from GloVe for English and those embedding from Tencent Embeddings for Chinese to extract the most similar 10 words (i.e., $m=10$) \footnote{The results of using other embeddings as sources to extract similar words are reported in the Appendix C.}.
% In addition, the number (i.e., $m$) of selected words carrying augmented semantic information is set to $10$. 
% 
In the augmentation module, we randomly initialize the embeddings of the extracted words (i.e., $\mathbf{e}_{i,j}$ for $c_{i,j}$) to represent the semantic information carried by those words.\footnote{We also try other ways (e.g., GloVe for English and Tencent Embedding for Chinese) to initialize the word embeddings, but do not find significant differences.}
% the similarity metrics for obtaining the similar words of each token is cosine similarity and
%

During the training process, we fix all pre-trained embeddings in the embedding layer and use Adam \cite{DBLP:journals/corr/KingmaB14} to optimize negative log-likelihood loss function with the learning rate set to $\eta=0.0001$, $\beta_1=0.9$ and $\beta_2=0.99$. 
We train 50 epochs for each method with the batch size set to $32$ and tune the hyper-parameters on the development set\footnote{We report the details of hyperparameter settings of different models in the Appendix D.}.
% , as well as their size and running speed in Appendix B.}}.
The model that achieves the best performance on the development set is evaluated on the test set with the $F1$ scores obtained from the official conlleval toolkits\footnote{The script to evaluate all models in the experiments is obtained from \url{https://www.clips.uantwerpen.be/conll2000/chunking/conlleval.txt}.}.

\subsection{Overall Results}

To explore the effect of the proposed attentive semantic augmentation module (${AU}$) and the gate module (${GA}$), we run different settings of our model with and without the modules.
In addition, we also try baselines that use direct summation ($DS$) to leverage the semantic information carried by the similar words, where the embeddings of the words are directly summed without weighting through attentions.
The experimental results ($F1$) of the baselines and our approach on the development and test sets of all datasets are reported in Table \ref{tab: results}(a) and (b), respectively.
% \footnote{The results of our models on the development sets are reported in Appendix B.}. 
% Firstly, to examine the effect of semantic augmentation, we run different methods, i.e. direct summation (sum the vectors for the augmented semantic information) and attention-based augmentation.

There are some observations from the results on the development and test sets.
First, compared to the baseline without semantic augmentation (ID=1), models using direct summation ($DS$, ID=2) to incorporate different semantic information undermines NER performance on two of three datasets, namely, W17 and WB;
on the contrary, the models using the proposed attentive semantic augmentation module ($AU$, ID=4) consistently outperform the baselines (ID=1 and ID=2) on 
% the development and test sets of 
all datasets.
It indicates that $AU$ could distinguish the contributions of different semantic information carried by different words in the given context and leverage them accordingly to improve NER performance.
Second, comparing the results of models with and without the gate module (${GA}$) (i.e. ID=3 vs. ID=2 and ID=5 vs. ID=4), we find that the models with gate module achieves superior performance to the others without it.
% on both development and test sets.
%
This observation suggests that the importance of the information from the context encoder and ${AU}$ varies, and the proposed gate module is effective in adjusting the weights according to their contributions.

\textcolor{black}{
Moreover, we compare our model under the best setting with previous models on all three datasets in Table \ref{tab:comparison}, where our model outperforms others on all datasets.
We believe that the new state-of-the-art performance is established.
The reason could be that compared to previous studies, our model is effective to alleviate the data sparsity problem in social media NER with the augmentation module to encode augmented semantic information.
Besides, the gate module can distinguish the importance of information from the context encoder and ${AU}$ according to their contribution to NER. 
}

\begin{table}[t]
    \centering
    % \small
    % \scalebox{0.9}{
    \begin{small}
    \begin{tabular}{l | C{1.0cm}C{1.0cm}C{1.0cm}}
    \toprule
    \textbf{Model} & \textbf{W16}  & \textbf{W17} & \textbf{WB} \\
    \midrule
    \newcite{DBLP:conf/acl/ZhangY18}                    & -                 & -                     & 58.79         \\
    \newcite{DBLP:journals/corr/abs-1911-04474}& 54.06                 & 48.98              & 65.03         \\
    \newcite{DBLP:conf/naacl/ZhuW19}                    & -                 & -                     & 59.31         \\
    \newcite{DBLP:conf/ijcai/GuiM0ZJH19} & - & - & 59.92 \\
    \newcite{DBLP:conf/emnlp/SuiCLZL19}                 & -                 & -                     & 63.09         \\
    \newcite{DBLP:conf/naacl/AkbikBV19}                         & -                 & 49.59 & - \\
    \newcite{DBLP:conf/acl/ZhouZJZFGK19}                        & 53.43                 & 42.83        & - \\
    \newcite{DBLP:conf/naacl/DevlinCLT19}       & 54.36 & 49.52 & 67.33 \\
    \newcite{DBLP:conf/nips/MengWWLNYLHSL19}    & - & - & 67.60 \\
    \newcite{DBLP:conf/cikm/XuWHL19}         & - & - & 68.93 \\
    \midrule
    Ours   & \textbf{55.01}& \textbf{50.36}& \textbf{69.80} \\
    \bottomrule
    \end{tabular}
    \end{small}
    % }53.41 / 52.16 / 68.04  and 50.06 / 48.97 / 65.74
    % }
    \vspace{-0.2cm}
    \caption{\textcolor{black}{Comparison of $F1$ scores of our best performing model (the full model with augmentation module and gate module) with that reported in previous studies on all English and Chinese social media datasets. }}
    \label{tab:comparison}
    \vskip -1.0em
\end{table}

\section{Analysis}

\subsection{Performance on Unseen Named Entities}

Since this work focuses on addressing the data sparsity problem in social media NER, where the unseen NEs
% \footnote{\textcolor{blue}{``Unseen NEs'' are the NEs in the test set that do not appear in the training set.}} 
are one of the important factors that hurts model performance.
To analyze whether our approach with attentive semantic augmentation ($AU$) and the gate module ($GA$) can address this problem, we report the recall of our approach (i.e., ``+$AU$+$GA$'') to recognize the unseen NEs on the test set of all datasets in Table \ref{tab:unseen}.
For reference, we also report the recall of the baseline without $AU$ and $GA$, as well as our runs of previous studies (marked by ``$^*$'').
% 
%
% 由于社交媒体文本的数据稀疏性问题，在inference阶段会遇到很多在训练阶段没有见过的NE，这也是其他模型在社交媒体领域表现差的原因。
% Due to the data sparsity of social media texts, many NEs encountered in the inference phase will not be seen in the training process, which is also the reason for the poor performance of previous models in social media domain. 
% 但是，我们的semantic augmentation模块可以解决这个问题。
% Nevertheless, our proposed model can alleviate this problem by leveraging rich augmented semantic information, encoding and aggregating such information by our proposed attention-based augmentation module and the gate module. 
% 为了证明我们模型的泛化性，我们将我们的best performing模型与没有使用augmentation的模型，以及previous模型对比，实验结果在table x中。
% For the sake of proving the generalization of our proposed model, we compare our best performing model (i.e. with $AU$ + $GA$) with the baseline model, and with previous approaches. 
% The experimental results ($recall$) are shown in Table \ref{tab:unseen}.
%我们可以观察到，我们的best performing model在unseen named entities上的表现比其他模型都要好，这说明我们的模型有更好的泛化性，可以更好的缓解数据稀疏性的问题。另外，我们带有SE的模型与没有SE的模型相比，可以看出这个能力的提升大多数是来自SE的贡献。
It is clearly observed that our approach outperforms the baseline and previous studies on unseen NEs on all datasets, which shows that it can appropriately leverage semantic information carried by similar words and thus alleviate the data sparsity problem. 
% In addition, the model with semantic augmentation performs better, compared with the variant without one, it can be seen that most of the improvement comes from the contribution of semantic augmentation.

\subsection{Case Study}

\begin{figure}[t]
    \centering
    \includegraphics[width=0.48\textwidth, trim=0 20 0 0]{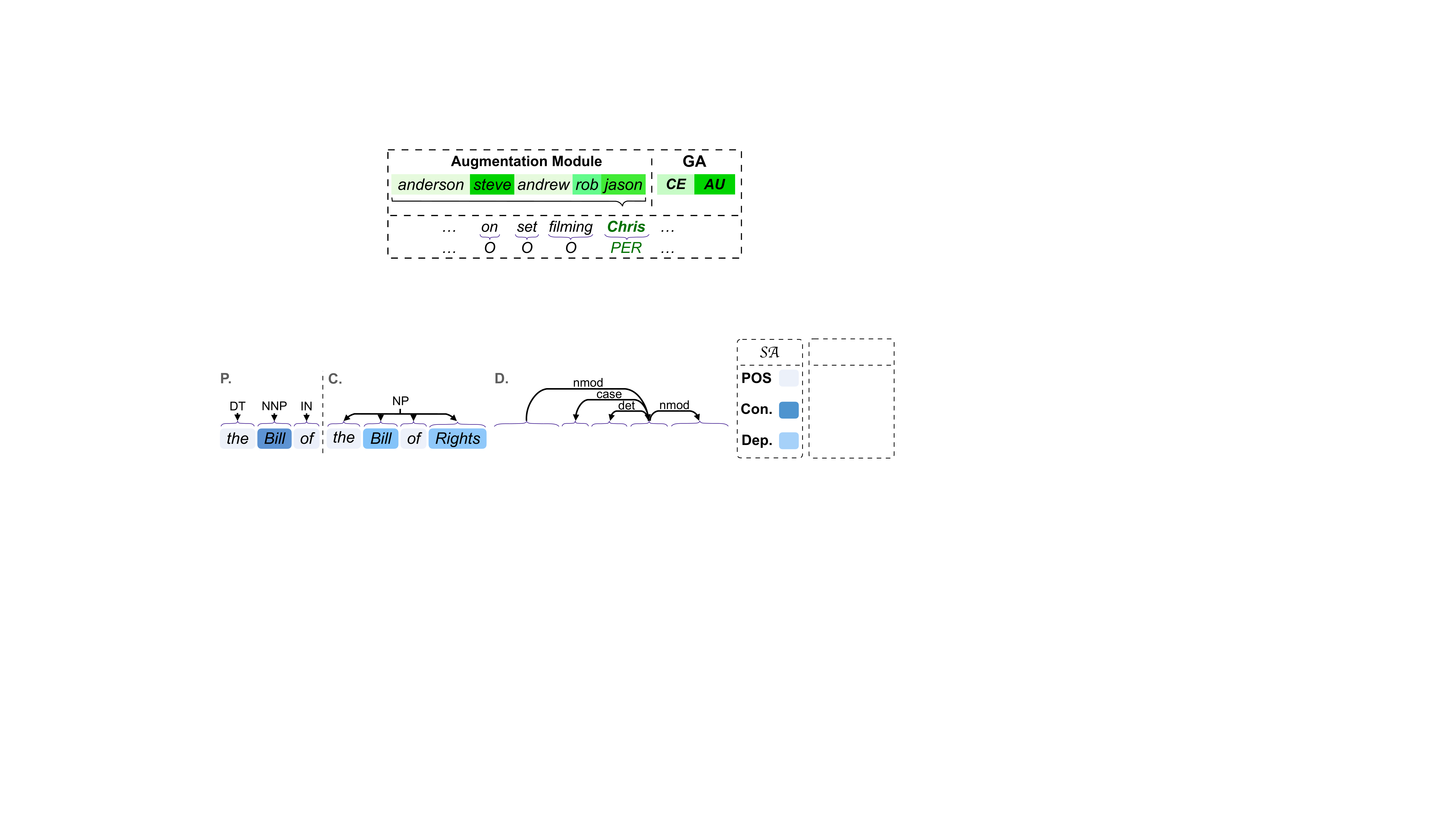}
    \caption{\textcolor{black}{An example of helping recognize the NE ``\textit{Chris}'' by augmented semantic information (darker color refers to greater value). ``$CE$'' and ``${AU}$'' represent the context encoder and attentive augmentation module, respectively. }}
    \label{fig:case}
    \vskip -1.0em
\end{figure}

% To better understand how the global context encoder module and fusion module model different type of information, in Figure X, we show the weights in global context encoder for identifying the importance of global context features and the weights in gate module to determine the information flow between local context information and global context information. In this case,
% To better understand how 
To demonstrate how the augmented semantic information improves NER with the attentive augmentation module and the gate module, 
% \textcolor{orange}{we present two cases in Figure \ref{fig:case} and Figure \ref{fig:case2}, 
we show the extracted augmented information for the word ``\textit{Chris}'' and visualize the weights for each augmented term in
Figure \ref{fig:case}, where deeper color refers to higher weight.
In this case, the words ``\textit{steve}'' and ``\textit{jason}'' have higher weights in ${AU}$.
The explanation could be that in all cases, these two words are a kind of ``\textit{Person}''. 
Thus, higher attention to these terms helps our model to identify the correct NE label. 
On the contrary, the term ``\textit{anderson}'' and ``\textit{andrew}'' never occur in the dataset, and therefore provide no helpful effect in this case and eventually end with the lower weights in ${AU}$. 
In addition, a model can also mislabel ``\textit{Chris}'' as ``\textit{Music-Artist}'', because ``\textit{Chris}'' belongs to that NE type in most cases and there is a word ``\textit{filming}'' in its context.
However, our model with the gate module can distinguish that the information from semantic augmentation is more important and thus make correct prediction.
% the context words in this sentence include ``\textit{filming}'', which may suggest a misleading ``\textit{Music-Artist}'' label.
% Thus, the augmented information is given more weights from ${GA}$.

% \textcolor{orange}{For the second case, the central word is ``\textit{anna}''. In the augmentation module, each term has appeared in the dataset and all refer to NE typed ``\textit{Person}''. So the weight distribution of these words is more even than the first case. However, the augmented information may mislead the model to give a ``\textit{Person}'' label. Therefore, the gate module gives more weight to $CE$, where the context information ``\textit{walk the moon}'' indicates that ``\textit{anna sun}'' may be a work of the band, and finally gives a correct label.}

\begin{table}[t]
    \centering
    % \small
    \scalebox{0.9}{
    % \begin{sc}
    \begin{tabular}{L{3.1cm}|C{1.25cm}C{1.25cm}C{1.1cm}}
        \toprule
        \textbf{Model} & \textbf{W16} & \textbf{W17} & \textbf{WB} \\
        \midrule
        \# of Unseen NEs & 2778 & 912 & 189 \\
        \midrule
        $^*$\citet{DBLP:conf/naacl/DevlinCLT19}       & 49.02            & 46.73            & 45.98 \\
        $^*$\citet{DBLP:journals/corr/abs-1911-04474} & 48.97            & 46.89            & 45.71 \\
        \midrule
        % \multicolumn{4}{l}{Ours} \\
        % \midrule
        % Ours & \\
        Baseline                                        & 49.04            & 46.72            & 45.79 \\
        Ours (+$AU$ +$GA$)                              & \textbf{51.27}   & \textbf{49.45}   & \textbf{48.81} \\
        \bottomrule
    \end{tabular}
    }
    % \end{sc}
    \vspace{-0.2cm}
    \caption{
    The recall of our models with and without the attentive semantic augmentation ($AU$) and the gate module ($GA$) on unseen named entities (whose numbers are also reported at the first row) on all three datasets. 
    The results of our runs of previous models (marked with ``$^*$'') are also reported for references.
    % $F1$ scores of the baseline model and ours enhanced with semantic augmentation (``${SE}$'') and the gate module (``${GA}$''). ``${DS}$'' and ``${AU}$'' represent the direct summation and attention-based augmentation module, respectively. $Y$ and $N$ denote the use and non-use of corresponding modules.
    }
    \label{tab:unseen}
    % \vskip -0.5em
\end{table}

% \begin{figure}[t]
%     \centering
%     \includegraphics[width=0.48\textwidth]{figure/figure_visualization_2.pdf}
%     \caption{\textcolor{orange}{
%     An example of helping recognize the NE ``\textit{anna}'' by leveraging the augmentation module and gating mechanism. 
%     }}
%     \label{fig:case2}
%     \vskip -1.0em
% \end{figure}

\section{Conclusion}
\textcolor{black}{
In this paper, we proposed a neural-based approach to enhance social media NER with semantic augmentation to alleviate data sparsity problem. 
Particularly, an 
% attention-based 
attentive semantic
augmentation module is suggested to encode semantic information and a gate module is applied to aggregate such information to tagging process. 
Experiments conducted on three benchmark datasets in English and Chinese show that our model outperforms previous studies and achieves the new state-of-the-art result. 
}

\bibliography{emnlp2020}

\begin{thebibliography}{41}
\expandafter\ifx\csname natexlab\endcsname\relax\def\natexlab#1{#1}\fi

\bibitem[{Aguilar et~al.(2017)Aguilar, Maharjan, L{\'{o}}pez{-}Monroy, and
  Solorio}]{DBLP:conf/aclnut/AguilarMLS17}
Gustavo Aguilar, Suraj Maharjan, Adri{\'{a}}n~Pastor L{\'{o}}pez{-}Monroy, and
  Thamar Solorio. 2017.
\newblock {A} {M}ulti-task {A}pproach for {N}amed {E}ntity {R}ecognition in
  {S}ocial {M}edia {D}ata.
\newblock In \emph{Proceedings of the 3rd Workshop on Noisy User-generated
  Text, NUT@EMNLP 2017, Copenhagen, Denmark, September 7, 2017}, pages
  148--153.

\bibitem[{Akbik et~al.(2019)Akbik, Bergmann, and
  Vollgraf}]{DBLP:conf/naacl/AkbikBV19}
Alan Akbik, Tanja Bergmann, and Roland Vollgraf. 2019.
\newblock {P}ooled {C}ontextualized {E}mbeddings for {N}amed {E}ntity
  {R}ecognition.
\newblock In \emph{Proceedings of the 2019 Conference of the North American
  Chapter of the Association for Computational Linguistics: Human Language
  Technologies, {NAACL-HLT} 2019, Minneapolis, MN, USA, June 2-7, 2019, Volume
  1 (Long and Short Papers)}, pages 724--728.

\bibitem[{Akbik et~al.(2018)Akbik, Blythe, and
  Vollgraf}]{DBLP:conf/coling/AkbikBV18}
Alan Akbik, Duncan Blythe, and Roland Vollgraf. 2018.
\newblock {C}ontextual {S}tring {E}mbeddings for {S}equence {L}abeling.
\newblock In \emph{Proceedings of the 27th International Conference on
  Computational Linguistics, {COLING} 2018, Santa Fe, New Mexico, USA, August
  20-26, 2018}, pages 1638--1649.

\bibitem[{Amjad et~al.(2020)Amjad, Sidorov, and Zhila}]{amjad-etal-2020-data}
Maaz Amjad, Grigori Sidorov, and Alisa Zhila. 2020.
\newblock Data {A}ugmentation using {M}achine {T}ranslation for {F}ake {N}ews
  {D}etection in the {U}rdu {U}language.
\newblock In \emph{Proceedings of the 12th Language Resources and Evaluation
  Conference}, pages 2537--2542, Marseille, France.

\bibitem[{Derczynski et~al.(2017)Derczynski, Nichols, van Erp, and
  Limsopatham}]{DBLP:conf/aclnut/DerczynskiNEL17}
Leon Derczynski, Eric Nichols, Marieke van Erp, and Nut Limsopatham. 2017.
\newblock {R}esults of the {WNUT2017} {S}hared {T}ask on {N}ovel and {E}merging
  {E}ntity {R}ecognition.
\newblock In \emph{Proceedings of the 3rd Workshop on Noisy User-generated
  Text, NUT@EMNLP 2017, Copenhagen, Denmark, September 7, 2017}, pages
  140--147.

\bibitem[{Devlin et~al.(2019)Devlin, Chang, Lee, and
  Toutanova}]{DBLP:conf/naacl/DevlinCLT19}
Jacob Devlin, Ming{-}Wei Chang, Kenton Lee, and Kristina Toutanova. 2019.
\newblock {BERT:} {P}re-training of {D}eep {B}idirectional {T}ransformers for
  {L}anguage {U}nderstanding.
\newblock In \emph{Proceedings of the 2019 Conference of the North American
  Chapter of the Association for Computational Linguistics: Human Language
  Technologies, {NAACL-HLT} 2019, Minneapolis, MN, USA, June 2-7, 2019, Volume
  1 (Long and Short Papers)}, pages 4171--4186.

\bibitem[{Diao et~al.(2019)Diao, Bai, Song, Zhang, and
  Wang}]{DBLP:journals/corr/zen}
Shizhe Diao, Jiaxin Bai, Yan Song, Tong Zhang, and Yonggang Wang. 2019.
\newblock {ZEN:} {P}re-training {C}hinese {T}ext {E}ncoder {E}nhanced by
  {N}-gram {R}epresentations.
\newblock \emph{Arxiv}, abs/1911.00720.

\bibitem[{Gui et~al.(2019)Gui, Ma, Zhang, Zhao, Jiang, and
  Huang}]{DBLP:conf/ijcai/GuiM0ZJH19}
Tao Gui, Ruotian Ma, Qi~Zhang, Lujun Zhao, Yu{-}Gang Jiang, and Xuanjing Huang.
  2019.
\newblock {CNN}-{B}ased {C}hinese {NER} with {L}exicon {R}ethinking.
\newblock In \emph{Proceedings of the Twenty-Eighth International Joint
  Conference on Artificial Intelligence, {IJCAI} 2019, Macao, China, August
  10-16, 2019}, pages 4982--4988.

\bibitem[{Jie and Lu(2019)}]{DBLP:conf/emnlp/JieL19}
Zhanming Jie and Wei Lu. 2019.
\newblock {D}ependency-{G}uided {LSTM-CRF} for {N}amed {E}ntity {R}ecognition.
\newblock In \emph{Proceedings of the 2019 Conference on Empirical Methods in
  Natural Language Processing and the 9th International Joint Conference on
  Natural Language Processing, {EMNLP-IJCNLP} 2019, Hong Kong, China, November
  3-7, 2019}, pages 3860--3870.

\bibitem[{Kasai et~al.(2019)Kasai, Friedman, Frank, Radev, and
  Rambow}]{DBLP:conf/naacl/KasaiFFRR19}
Jungo Kasai, Dan Friedman, Robert Frank, Dragomir~R. Radev, and Owen Rambow.
  2019.
\newblock {S}yntax-aware {N}eural {S}emantic {R}ole {L}abeling with
  {S}upertags.
\newblock In \emph{Proceedings of the 2019 Conference of the North American
  Chapter of the Association for Computational Linguistics: Human Language
  Technologies, {NAACL-HLT} 2019, Minneapolis, MN, USA, June 2-7, 2019, Volume
  1 (Long and Short Papers)}, pages 701--709.

\bibitem[{Kim et~al.(2019)Kim, Kang, and Kwak}]{DBLP:conf/aaai/KimKK19}
Seonhoon Kim, Inho Kang, and Nojun Kwak. 2019.
\newblock {S}emantic {S}entence {M}atching with {D}ensely-{C}onnected
  {R}ecurrent and {C}o-{A}ttentive {I}nformation.
\newblock In \emph{The Thirty-Third {AAAI} Conference on Artificial
  Intelligence, {AAAI} 2019, The Thirty-First Innovative Applications of
  Artificial Intelligence Conference, {IAAI} 2019, The Ninth {AAAI} Symposium
  on Educational Advances in Artificial Intelligence, {EAAI} 2019, Honolulu,
  Hawaii, USA, January 27 - February 1, 2019}, pages 6586--6593.

\bibitem[{Kingma and Ba(2015)}]{DBLP:journals/corr/KingmaB14}
Diederik~P. Kingma and Jimmy Ba. 2015.
\newblock {A}dam: {A} {M}ethod for {S}tochastic {O}ptimization.
\newblock In \emph{3rd International Conference on Learning Representations,
  {ICLR} 2015, San Diego, CA, USA, May 7-9, 2015, Conference Track
  Proceedings}.

\bibitem[{Kumar et~al.(2018)Kumar, Kawahara, and
  Kurohashi}]{kumar-etal-2018-knowledge}
Abhishek Kumar, Daisuke Kawahara, and Sadao Kurohashi. 2018.
\newblock {K}nowledge-{E}nriched {T}wo-{L}ayered {A}ttention {N}etwork for
  {S}entiment {A}nalysis.
\newblock In \emph{Proceedings of the 2018 Conference of the North {A}merican
  Chapter of the Association for Computational Linguistics: Human Language
  Technologies, Volume 2 (Short Papers)}, pages 253--258, New Orleans,
  Louisiana.

\bibitem[{Kumar et~al.(2019)Kumar, Glaude, de~Lichy, and
  Campbell}]{kumar-etal-2019-closer}
Varun Kumar, Hadrien Glaude, Cyprien de~Lichy, and Wlliam Campbell. 2019.
\newblock A {C}loser {L}ook at {F}eature {S}pace {D}ata {A}ugmentation for
  {F}ew-{S}hot {I}ntent {C}lassification.
\newblock In \emph{Proceedings of the 2nd Workshop on Deep Learning Approaches
  for Low-Resource NLP (DeepLo 2019)}, pages 1--10, Hong Kong, China.

\bibitem[{Lample et~al.(2016)Lample, Ballesteros, Subramanian, Kawakami, and
  Dyer}]{DBLP:conf/naacl/LampleBSKD16}
Guillaume Lample, Miguel Ballesteros, Sandeep Subramanian, Kazuya Kawakami, and
  Chris Dyer. 2016.
\newblock {N}eural {A}rchitectures for {N}amed {E}ntity {R}ecognition.
\newblock In \emph{{NAACL} {HLT} 2016, The 2016 Conference of the North
  American Chapter of the Association for Computational Linguistics: Human
  Language Technologies, San Diego California, USA, June 12-17, 2016}, pages
  260--270.

\bibitem[{Manning et~al.(2014)Manning, Surdeanu, Bauer, Finkel, Bethard, and
  McClosky}]{DBLP:conf/acl/ManningSBFBM14}
Christopher~D. Manning, Mihai Surdeanu, John Bauer, Jenny~Rose Finkel, Steven
  Bethard, and David McClosky. 2014.
\newblock {T}he {S}tanford {C}orenlp {N}atural {L}anguage {P}rocessing
  {T}oolkit.
\newblock In \emph{Proceedings of the 52nd Annual Meeting of the Association
  for Computational Linguistics, {ACL} 2014, June 22-27, 2014, Baltimore, MD,
  USA, System Demonstrations}, pages 55--60.

\bibitem[{Margatina et~al.(2019)Margatina, Baziotis, and
  Potamianos}]{margatina-etal-2019-attention}
Katerina Margatina, Christos Baziotis, and Alexandros Potamianos. 2019.
\newblock {A}ttention-based {C}onditioning {M}ethods for {E}xternal {K}nowledge
  {I}ntegration.
\newblock In \emph{Proceedings of the 57th Annual Meeting of the Association
  for Computational Linguistics}, pages 3944--3951, Florence, Italy.

\bibitem[{Martins et~al.(2019)Martins, Marinho, and
  Martins}]{DBLP:conf/acl/MartinsMM19}
Pedro~Henrique Martins, Zita Marinho, and Andr{\'{e}} F.~T. Martins. 2019.
\newblock {J}oint {L}earning of {N}amed {E}ntity {R}ecognition and {E}ntity
  {L}inking.
\newblock In \emph{Proceedings of the 57th Conference of the Association for
  Computational Linguistics, {ACL} 2019, Florence, Italy, July 28 - August 2,
  2019, Volume 2: Student Research Workshop}, pages 190--196.

\bibitem[{Meng et~al.(2019)Meng, Wu, Wang, Li, Nie, Yin, Li, Han, Sun, and
  Li}]{DBLP:conf/nips/MengWWLNYLHSL19}
Yuxian Meng, Wei Wu, Fei Wang, Xiaoya Li, Ping Nie, Fan Yin, Muyu Li, Qinghong
  Han, Xiaofei Sun, and Jiwei Li. 2019.
\newblock {G}lyce: {G}lyph-vectors for {C}hinese {C}haracter {R}epresentations.
\newblock In \emph{Advances in Neural Information Processing Systems 32: Annual
  Conference on Neural Information Processing Systems 2019, NeurIPS 2019, 8-14
  December 2019, Vancouver, BC, Canada}, pages 2742--2753.

\bibitem[{Mikolov et~al.(2013)Mikolov, Chen, Corrado, and
  Dean}]{DBLP:journals/corr/word2vec}
Tomas Mikolov, Kai Chen, Greg Corrado, and Jeffrey Dean. 2013.
\newblock {E}fficient {E}stimation of {W}ord {R}epresentations in {V}ector
  {S}pace.
\newblock In \emph{1st International Conference on Learning Representations,
  {ICLR} 2013, Scottsdale, Arizona, USA, May 2-4, 2013, Workshop Track
  Proceedings}.

\bibitem[{Pang et~al.(2019)Pang, Lan, Guo, Xu, Su, and
  Cheng}]{DBLP:conf/aaai/PangLGXSC19}
Liang Pang, Yanyan Lan, Jiafeng Guo, Jun Xu, Lixin Su, and Xueqi Cheng. 2019.
\newblock {HAS-QA:} {H}ierarchical {A}nswer {S}pans {M}odel for {O}pen-{D}omain
  {Q}uestion {A}nswering.
\newblock In \emph{The Thirty-Third {AAAI} Conference on Artificial
  Intelligence, {AAAI} 2019, The Thirty-First Innovative Applications of
  Artificial Intelligence Conference, {IAAI} 2019, The Ninth {AAAI} Symposium
  on Educational Advances in Artificial Intelligence, {EAAI} 2019, Honolulu,
  Hawaii, USA, January 27 - February 1, 2019}, pages 6875--6882.

\bibitem[{Peng and Dredze(2015)}]{DBLP:conf/emnlp/PengD15}
Nanyun Peng and Mark Dredze. 2015.
\newblock {N}amed {E}ntity {R}ecognition for {C}hinese {S}ocial {M}edia with
  {J}ointly {T}rained {E}mbeddings.
\newblock In \emph{Proceedings of the 2015 Conference on Empirical Methods in
  Natural Language Processing, {EMNLP} 2015, Lisbon, Portugal, September 17-21,
  2015}, pages 548--554.

\bibitem[{Pennington et~al.(2014)Pennington, Socher, and
  Manning}]{DBLP:conf/emnlp/PenningtonSM14}
Jeffrey Pennington, Richard Socher, and Christopher~D. Manning. 2014.
\newblock {G}love: {G}lobal {V}ectors for {W}ord {R}epresentation.
\newblock In \emph{Proceedings of the 2014 Conference on Empirical Methods in
  Natural Language Processing, {EMNLP} 2014, October 25-29, 2014, Doha, Qatar,
  {A} meeting of SIGDAT, a Special Interest Group of the {ACL}}, pages
  1532--1543.

\bibitem[{Peters et~al.(2018)Peters, Neumann, Iyyer, Gardner, Clark, Lee, and
  Zettlemoyer}]{DBLP:conf/naacl/PetersNIGCLZ18}
Matthew~E. Peters, Mark Neumann, Mohit Iyyer, Matt Gardner, Christopher Clark,
  Kenton Lee, and Luke Zettlemoyer. 2018.
\newblock {D}eep {C}ontextualized {W}ord {R}epresentations.
\newblock In \emph{Proceedings of the 2018 Conference of the North American
  Chapter of the Association for Computational Linguistics: Human Language
  Technologies, {NAACL-HLT} 2018, New Orleans, Louisiana, USA, June 1-6, 2018,
  Volume 1 (Long Papers)}, pages 2227--2237.

\bibitem[{Ritter et~al.(2011)Ritter, Clark, Mausam, and
  Etzioni}]{DBLP:conf/emnlp/RitterCME11}
Alan Ritter, Sam Clark, Mausam, and Oren Etzioni. 2011.
\newblock {N}amed {E}ntity {R}ecognition in {T}weets: {A}n {E}xperimental
  {S}tudy.
\newblock In \emph{Proceedings of the 2011 Conference on Empirical Methods in
  Natural Language Processing, {EMNLP} 2011, 27-31 July 2011, John McIntyre
  Conference Centre, Edinburgh, UK, {A} meeting of SIGDAT, a Special Interest
  Group of the {ACL}}, pages 1524--1534.

\bibitem[{Sileo et~al.(2019)Sileo, Van De~Cruys, Pradel, and
  Muller}]{sileo-etal-2019-mining}
Damien Sileo, Tim Van De~Cruys, Camille Pradel, and Philippe Muller. 2019.
\newblock {M}ining {D}iscourse {M}arkers for {U}nsupervised {S}entence
  {R}epresentation {L}earning.
\newblock In \emph{Proceedings of the 2019 Conference of the North {A}merican
  Chapter of the Association for Computational Linguistics: Human Language
  Technologies, Volume 1 (Long and Short Papers)}, pages 3477--3486,
  Minneapolis, Minnesota.

\bibitem[{Song et~al.(2017)Song, Lee, and Xia}]{song2017learning}
Yan Song, Chia-Jung Lee, and Fei Xia. 2017.
\newblock Learning {W}ord {R}epresentations with {R}egularization from {P}rior
  {K}nowledge.
\newblock In \emph{Proceedings of the 21st Conference on Computational Natural
  Language Learning (CoNLL 2017)}, pages 143--152.

\bibitem[{Song et~al.(2018{\natexlab{a}})Song, Shi, and
  Li}]{song-etal-2018-joint}
Yan Song, Shuming Shi, and Jing Li. 2018{\natexlab{a}}.
\newblock Joint {L}earning {E}mbeddings for {C}hinese {W}ords and their
  {C}omponents via {L}adder {S}tructured {N}etworks.
\newblock In \emph{Proceedings of the Twenty-Seventh International Joint
  Conference on Artificial Intelligence, {IJCAI-18}}, pages 4375--4381.

\bibitem[{Song et~al.(2018{\natexlab{b}})Song, Shi, Li, and
  Zhang}]{DBLP:conf/naacl/SongSLZ18}
Yan Song, Shuming Shi, Jing Li, and Haisong Zhang. 2018{\natexlab{b}}.
\newblock {D}irectional {S}kip-{G}ram: {E}xplicitly {D}istinguishing {L}eft and
  {R}ight {C}ontext for {W}ord {E}mbeddings.
\newblock In \emph{Proceedings of the 2018 Conference of the North American
  Chapter of the Association for Computational Linguistics: Human Language
  Technologies, NAACL-HLT, New Orleans, Louisiana, USA, June 1-6, 2018, Volume
  2 (Short Papers)}, pages 175--180.

\bibitem[{Song and Xia(2013)}]{song-xia-2013-common}
Yan Song and Fei Xia. 2013.
\newblock A {C}ommon {C}ase of {J}ekyll and {H}yde: {T}he {S}ynergistic
  {E}ffect of {U}sing {D}ivided {S}ource {T}raining {D}ata for {F}eature
  {A}ugmentation.
\newblock In \emph{Proceedings of the Sixth International Joint Conference on
  Natural Language Processing}, pages 623--631, Nagoya, Japan.

\bibitem[{Strauss et~al.(2016)Strauss, Toma, Ritter, de~Marneffe, and
  Xu}]{DBLP:conf/aclnut/StraussTRMX16}
Benjamin Strauss, Bethany Toma, Alan Ritter, Marie{-}Catherine de~Marneffe, and
  Wei Xu. 2016.
\newblock {R}esults of the {WNUT16} {N}amed {E}ntity {R}ecognition {S}hared
  {T}ask.
\newblock In \emph{Proceedings of the 2nd Workshop on Noisy User-generated
  Text, NUT@COLING 2016, Osaka, Japan, December 11, 2016}, pages 138--144.

\bibitem[{Sui et~al.(2019)Sui, Chen, Liu, Zhao, and
  Liu}]{DBLP:conf/emnlp/SuiCLZL19}
Dianbo Sui, Yubo Chen, Kang Liu, Jun Zhao, and Shengping Liu. 2019.
\newblock {L}everage {L}exical {K}nowledge for {C}hinese {N}amed {E}ntity
  {R}ecognition via {C}ollaborative {G}raph {N}etwork.
\newblock In \emph{Proceedings of the 2019 Conference on Empirical Methods in
  Natural Language Processing and the 9th International Joint Conference on
  Natural Language Processing, {EMNLP-IJCNLP} 2019, Hong Kong, China, November
  3-7, 2019}, pages 3828--3838.

\bibitem[{Tian et~al.(2020{\natexlab{a}})Tian, Song, Ao, Xia, Quan, Zhang, and
  Wang}]{tian-etal-2020-joint}
Yuanhe Tian, Yan Song, Xiang Ao, Fei Xia, Xiaojun Quan, Tong Zhang, and
  Yonggang Wang. 2020{\natexlab{a}}.
\newblock {J}oint {C}hinese {W}ord {S}egmentation and {P}art-of-speech
  {T}agging via {T}wo-way {A}ttentions of {A}uto-analyzed {K}nowledge.
\newblock In \emph{Proceedings of the 58th Annual Meeting of the Association
  for Computational Linguistics}, pages 8286--8296, Online.

\bibitem[{Tian et~al.(2020{\natexlab{b}})Tian, Song, and
  Xia}]{tian-etal-2020-suppertagging}
Yuanhe Tian, Yan Song, and Fei Xia. 2020{\natexlab{b}}.
\newblock Supertagging {C}ombinatory {C}ategorial {G}rammar with {A}ttentive
  {G}raph {C}onvolutional {N}etworks.
\newblock In \emph{Proceedings of the 2020 Conference on Empirical Methods in
  Natural Language Processing}.

\bibitem[{Tian et~al.(2020{\natexlab{c}})Tian, Song, Xia, and
  Zhang}]{tian-etal-2020-constituency}
Yuanhe Tian, Yan Song, Fei Xia, and Tong Zhang. 2020{\natexlab{c}}.
\newblock Improving {C}onstituency {P}arsing with {S}pan {A}ttention.
\newblock In \emph{Findings of the 2020 Conference on Empirical Methods in
  Natural Language Processing}.

\bibitem[{Tian et~al.(2020{\natexlab{d}})Tian, Song, Xia, Zhang, and
  Wang}]{tian2020improving}
Yuanhe Tian, Yan Song, Fei Xia, Tong Zhang, and Yonggang Wang.
  2020{\natexlab{d}}.
\newblock {I}mproving {C}hinese {W}ord {S}egmentation with {W}ordhood {M}emory
  {N}etworks.
\newblock In \emph{Proceedings of the 58th Annual Meeting of the Association
  for Computational Linguistics}, pages 8274--8285.

\bibitem[{Xu et~al.(2019)Xu, Wang, Han, and Li}]{DBLP:conf/cikm/XuWHL19}
Canwen Xu, Feiyang Wang, Jialong Han, and Chenliang Li. 2019.
\newblock {E}xploiting {M}ultiple {E}mbeddings for {C}hinese {N}amed {E}ntity
  {R}ecognition.
\newblock In \emph{Proceedings of the 28th {ACM} International Conference on
  Information and Knowledge Management, {CIKM} 2019, Beijing, China, November
  3-7, 2019}, pages 2269--2272.

\bibitem[{Yan et~al.(2019)Yan, Deng, Li, and
  Qiu}]{DBLP:journals/corr/abs-1911-04474}
Hang Yan, Bocao Deng, Xiaonan Li, and Xipeng Qiu. 2019.
\newblock \href {http://arxiv.org/abs/1911.04474} {{TENER:} {A}dapting
  {T}ransformer {E}ncoder for {N}amed {E}ntity {R}ecognition}.
\newblock \emph{arXiv}, abs/1911.04474.

\bibitem[{Zhang and Yang(2018)}]{DBLP:conf/acl/ZhangY18}
Yue Zhang and Jie Yang. 2018.
\newblock {C}hinese {NER} {U}sing {L}attice {LSTM}.
\newblock In \emph{Proceedings of the 56th Annual Meeting of the Association
  for Computational Linguistics, {ACL} 2018, Melbourne, Australia, July 15-20,
  2018, Volume 1: Long Papers}, pages 1554--1564.

\bibitem[{Zhou et~al.(2019)Zhou, Zhang, Jin, Zhu, Fang, Goh, and
  Kwok}]{DBLP:conf/acl/ZhouZJZFGK19}
Joey~Tianyi Zhou, Hao Zhang, Di~Jin, Hongyuan Zhu, Meng Fang, Rick Siow~Mong
  Goh, and Kenneth Kwok. 2019.
\newblock {D}ual {A}dversarial {N}eural {T}ransfer for {L}ow-{R}esource {N}amed
  {E}ntity {R}ecognition.
\newblock In \emph{Proceedings of the 57th Conference of the Association for
  Computational Linguistics, {ACL} 2019, Florence, Italy, July 28- August 2,
  2019, Volume 1: Long Papers}, pages 3461--3471.

\bibitem[{Zhu and Wang(2019)}]{DBLP:conf/naacl/ZhuW19}
Yuying Zhu and Guoxin Wang. 2019.
\newblock {CAN-NER:} {C}onvolutional {A}ttention {N}etwork for {C}hinese
  {N}amed {E}ntity {R}ecognition.
\newblock In \emph{Proceedings of the 2019 Conference of the North American
  Chapter of the Association for Computational Linguistics: Human Language
  Technologies, {NAACL-HLT} 2019, Minneapolis, MN, USA, June 2-7, 2019, Volume
  1 (Long and Short Papers)}, pages 3384--3393.

\end{thebibliography}
\bibliographystyle{acl_natbib}

\vspace{0.4cm}

% \clearpage

\appendix

\section*{Appendix A: Effect of Using Different Embeddings} \label{app: embedding}

\vspace{-0.2cm}

\begin{table}[h]
    \centering
    \small
    % \scalebox{0.85}{
    \begin{tabular}{l L{0.9cm} | C{1.1cm}C{1.1cm}C{0.9cm}}
        \toprule
        \textbf{Model} & \textbf{Emb.} & \textbf{W16} & \textbf{W17} & \textbf{WB} \\
        \midrule
        Baseline            & \multirow{2}{*}{ELMo} & 52.16             & 47.31             & -             \\
        \ \ +$AU$+$GA$   &                          & 54.31             & 48.76             & -             \\
        \cmidrule{1-5}
        Baseline            & \multirow{2}{*}{BERT} & 52.09             & 48.33             & -             \\
        \ \ +$AU$+$GA$   &                          & \textbf{54.16}    & \textbf{49.57}    & -             \\
        \midrule
        Baseline            & \multirow{2}{*}{Tencent}& -                 & -               & 60.54              \\
        \ \ +$AU$+$GA$   &                          & -                 & -                 & 63.12         \\
        \cmidrule{1-5}
        Baseline            & \multirow{2}{*}{ZEN}   & -                 & -                 & 66.09              \\
        \ \ +$AU$+$GA$   &                          & -                 & -                 & \textbf{68.96}\\
        \bottomrule
    \end{tabular}
    % }
    \vspace{-0.2cm}
    \caption{
    % The hyper-parameters and hyper-parameter search ranges for our models.
    Experimental results ($F1$ scores) of our approach with semantic augmentation ($AU$) and gate module ($GA$) on all datasets, where only one type of embeddings is used in the embedding layer to represent the input sentence.
    The results of their corresponding baseline without $AU$ and $GA$ are also reported.
    % Experimental results ($F1$ scores) of our best performing model (i.e., the full model with $AU$ and $GA$) using different pre-trained embeddings as the source of getting augmented semantic information
    %search ranges for our models and the best parameters for all datasets.
    }
    \label{tab:embedding}
    \vskip -1.0em
\end{table}

% \begin{table}[t]
%     \centering
%     \small
    
%     \begin{subtable}[t]{0.48\textwidth}
%     \centering
%     \small
%     \begin{tabular}{L{2.3cm}C{1.3cm}C{1.15cm}C{1.15cm}}
%     \toprule
%     \textbf{Model} & \textbf{Embedding} & \textbf{WNUT16} & \textbf{WNUT17} \\
%     \midrule
%     Baseline    & ELMo  & & \\
%     Ours (+$AU$+$GA$) & ELMo & & \\
%     Baseline    & BERT  & & \\
%     Ours (+$AU$+$GA$) & BERT & & \\
%     \bottomrule
%     \end{tabular}
%     \caption{\textcolor{orange}{The English datasets. }}
%     \vspace{0.3cm}
%     \end{subtable}
    
%     \begin{subtable}[t]{0.48\textwidth}
%     \centering
%     \small
%     \begin{tabular}{L{2.3cm}C{1.3cm}C{2.3cm}}
%     \toprule
%     \textbf{Model} & \textbf{Embedding} & \textbf{Weibo} \\
%     \midrule
%     Baseline    & Tencent  & \\
%     Ours (+$AU$+$GA$) & Tencent & \\
%     Baseline    & ZEN  & \\
%     Ours (+$AU$+$GA$) & ZEN & \\
%     \bottomrule
%     \end{tabular}
%     \caption{\textcolor{orange}{The Chinese dataset. }}
%     % \vspace{cm}
%     \end{subtable}
%     \caption{\textcolor{orange}{Caption. }}
%     \label{tab:embedding}
% \end{table}

In our main experiments, we use two types of embeddings for each language: ELMo \cite{DBLP:conf/naacl/PetersNIGCLZ18} and BERT-cased large  \cite{DBLP:conf/naacl/DevlinCLT19} for English, and Tencent Embedding \cite{DBLP:conf/naacl/SongSLZ18} and ZEN \cite{DBLP:journals/corr/zen} for Chinese.
In Table \ref{tab:embedding}, we report the results ($F1$ scores) of our model with the best setting (i.e. the full model with semantic augmentation ($AU$) and gate module ($GA$)) as well as the baselines without $AU$ and $GA$, where either one of the two types of embedding is used to represent the input sentence.
From the results, it is found that our model with $AU$ and $GA$ can consistently outperforms the baseline models with different settings of embeddings.
%
% \textcolor{orange}{
% Since different types of embeddings carry contextual information learned from different corpora and algorithms, we try different pretrained embeddings and their pairwise combination. 
% The experiment is performed on our best setting (i.e. the full model with $AU$ and $GA$), with the results ($F1$ scores) reported in Table \ref{tab:embedding}. 
% }

\section*{Appendix B: Comparison Between BERT and ZEN on WB}

\vspace{-0.2cm}
\begin{table}[h]
    \centering
    \begin{tabular}{L{3.8cm} | C{2.4cm}}
        \toprule
        \textbf{Embeddings} & \textbf{WB}\\
        \midrule
        BERT + Tencent      & 69.56         \\
        ZEN + Tencent       & \textbf{69.80}\\
        \bottomrule
    \end{tabular}
    \vspace{-0.2cm}
    \caption{Experimental results ($F1$ scores) of our model with $AU$ and $GA$ on the WB dataset, where BERT or ZEN is used as one of the two types of embeddings (the other one is Tencent Embedding) to represent the input sentence for the embedding layer.}
    \label{tab:bert}
    \vskip -1em
\end{table}

\noindent
In our main experiments, we use ZEN \cite{DBLP:journals/corr/zen} instead of BERT \cite{DBLP:conf/naacl/DevlinCLT19} as the embedding to represent the input for Chinese.
The reason is that ZEN achieves better performance compared with BERT, which is confirmed by
Table \ref{tab:bert} with its results ($F1$ scores) showing the performance of our approach with the best settings (i.e. two types of embeddings with $AU$ and $GA$.) on WB dataset. Either BERT or ZEN is used as one of the two types of embeddings (the other type of embedding is Tencent Embedding).
%
%The results shows that our model with ZEN outperform the one with BERT.

\section*{Appendix C: Effect of Using Different Embeddings to Extract Similar Words} \label{app: source}

\vspace{-0.2cm}
\begin{table}[h]
    \centering
    \small
    % \scalebox{0.85}{
    \begin{tabular}{l L{1.5cm} | C{0.8cm}C{0.8cm}C{0.8cm}}
        \toprule
        \textbf{Model} & \textbf{Source} & \textbf{W16} & \textbf{W17} & \textbf{WB} \\
        \midrule
        \multicolumn{2}{l|}{Baseline} & 49.56             & 49.11             & 66.02             \\
        \midrule
           & Word2vec  & 54.94             & 50.22             & -             \\
        Ours  & GloVe     & \textbf{55.01}    & \textbf{50.36}    & -             \\
        \addlinespace[0.05cm]
        \cline{2-5}
        \addlinespace[0.05cm]
        (+$AU$+$GA$)   & Giga      & -                 & -                 & 69.68         \\
           & Tencent   & -                 & -                 & \textbf{69.80}\\
        \bottomrule
    \end{tabular}
    % }
    \vspace{-0.3cm}
    \caption{
    % The hyper-parameters and hyper-parameter search ranges for our models. 
    Experimental results ($F1$ scores) of our best performing models (i.e., the ones with $AU$ and $GA$) using different types of pre-trained embeddings as the source to extract similar words.
    The results of baseline (the one without $AU$ and $GA$) are also reported.}
    %search ranges for our models and the best parameters for all datasets.
    \label{tab:source}
    \vskip -1em
\end{table}

In addition to use embeddings for input sentence representation,
we also try different embedding sources (i.e. pre-trained word embeddings) to extract similar words for each token in the input sentence.
For English, we use Word2vec \cite{DBLP:journals/corr/word2vec} and Glove \cite{DBLP:conf/acl/ManningSBFBM14}; for Chinese, we use Giga \cite{DBLP:conf/acl/ZhangY18} and Tencent Embedding \cite{DBLP:conf/naacl/SongSLZ18}.\footnote{We obtain Word2vec from \url{https://code.google.com/archive/p/word2vec/}, GloVe from \url{https://nlp.stanford.edu/projects/glove/}, Giga from \url{https://github.com/jiesutd/LatticeLSTM}.}
%, and Tencent Embedding from \url{https://ai.tencent.com/ailab/nlp/embedding.html}.}
% 
% To extract similar words carrying augmented semantic information, we try different pre-trained word embeddings as sources to extract such information. 
The experimental results of our model with the best setting (i.e., the one with $AU$ and $GA$) using different sources are reported in Table \ref{tab:source}.
The result of the baseline model without $AU$ and $GA$ is also reported for reference.
The results show that our approach can consistently outperforms the baseline with different sources to find similar words, which demonstrates the robustness of our approach.
% We find that Glove achieves better performance than Word2Vec for English and Tencent Embedding achieves better performance than Giga

% \vspace{0.2cm}

\section*{Appendix D: Hyper-parameter Settings}
\label{app: hyper-para}

\vspace{-0.2cm}
\begin{table}[th]
    \centering
    \small
    % \scalebox{0.95}{
    \begin{tabular}{l | l | r}
        \toprule
        % \textbf{Hyper-parameter Types} & \textbf{Hyper-Parameter Search Ranges} & \multirow{2}{*}{\textbf{Best}} \\
         & \textbf{Values} & \textbf{Best} \\
        \midrule
        Dropout rate            & $0$, $0.1$, $0.2$, $0.3$      & $0.2$ \\
        Learning rate           & $e^{-5}$, $e^{-4}$, $e^{-3}$  & $e^{-4}$  \\
        % $\beta_1$ of Adam         & $0.9$                         & $0.9$ \\
        % $\beta_2$ of Adam         & $0.99$                        & $0.99$    \\
        Batch size              & $8$, $16$, $32$               & $32$ \\
        Number of layers        & $1$, $2$, $4$                 & $2$ \\
        Number of head          & $4$, $8$, $12$                & $12$ \\
        Hidden units            & $64$, $128$, $256$            & $128$ \\
        \# of similar of words ($m$)                  &  $5$, $10$, $20$              & $10$ \\  
        \bottomrule
    \end{tabular}
    % }
    \vspace{-0.2cm}
    \caption{
    All values of different hyper-parameters as well as the best one used in our experiments.
    }
    \label{tab:hyper}
    \vskip -1em
\end{table}

\noindent
We report all values of the hyper-parameters tried for our models in Table \ref{tab:hyper},
where
we try different combinations of them and find the best hyper-parameter configurations (which is also reported in Table \ref{tab:hyper}) on the development set of each dataset.
% The best values for all hyperparameters are also reported, which are obtained by tuning our model with the given hyperparameter values on the development set of each dataset.

\end{document}